\documentclass[runningheads]{llncs}

\usepackage{eccv}

\usepackage{eccvabbrv}

\usepackage{graphicx}
\usepackage{booktabs}

\usepackage[accsupp]{axessibility}  % Improves PDF readability for those with disabilities.
\newcommand{\PAR}[1]{\vspace{-0.2eM}\vskip4pt \noindent{\bf #1}}

\usepackage[pagebackref,breaklinks,colorlinks,citecolor=eccvblue]{hyperref}
\usepackage{orcidlink}

\usepackage[dvipsnames]{xcolor}
\usepackage{graphicx}
\usepackage{amsmath}
\usepackage{amssymb}
\usepackage{multicol, blindtext}
\usepackage{subcaption}
\usepackage{cite}
\usepackage{soul}
\usepackage{units}
\usepackage{array}
\usepackage{multirow}
\usepackage{caption}
\usepackage{float}
\usepackage{booktabs}
\usepackage{tabularx}
\usepackage{nicefrac}
\usepackage{enumitem}
\usepackage{adjustbox}
\usepackage{adjustbox}

\newcolumntype{R}[2]{%
    >{\adjustbox{angle=#1,lap=\width-(#2)}\bgroup}%
    l%
    <{\egroup}%
}

\newcolumntype{L}[1]{>{\raggedright\arraybackslash}p{#1}}
\newcolumntype{C}[1]{>{\centering\arraybackslash}p{#1}}
\newcolumntype{R}[1]{>{\raggedleft\arraybackslash}p{#1}}
\usepackage{pifont}
\newcommand{\cmark}{\ding{51}}
\newcommand{\xmark}{\ding{55}}

\usepackage{caption}
\usepackage{xcolor}
\usepackage{mathtools}
\usepackage{xfrac}
\usepackage{wasysym}
\usepackage{algorithm}% http://ctan.org/pkg/algorithm
\usepackage{algpseudocode}% http://ctan.org/pkg/algorithmicx
\usepackage[accsupp]{axessibility}  % Improves PDF readability for those with disabilities

\begin{document}

\title{SAMFusion: Sensor-Adaptive Multimodal Fusion for 3D Object Detection in Adverse Weather}

\titlerunning{SAMFusion: Sensor-Adaptive Multimodal Fusion Method}

\author{
Edoardo Palladin$^*$ \inst{1}\orcidlink{0009-0005-2948-5368}  \and
Roland Dietze$^{*}$\inst{2}\orcidlink{0009-0006-1337-0205} \and
Praveen Narayanan \inst{1}\orcidlink{0009-0007-7097-6227} \and \\
Mario Bijelic \inst{1,3}\orcidlink{0000-0002-2676-9833} \and
Felix Heide \inst{1,3}\orcidlink{0000-0002-8054-9823}
}

\authorrunning{Palladin \and Dietze et al.}

% TODO FINAL: Replace with your institution list

\institute{
 $^{1}$ Torc Robotics \quad
 $^{2}$ University of Stuttgart \quad
 $^{3}$ Princeton University
}

\maketitle

\def\thefootnote{$*$}\footnotetext{These authors contributed equally to this work.} 
\def\thefootnote{$1$}\footnotetext{https://light.princeton.edu/samfusion/}

\begin{abstract}
Multimodal sensor fusion is an essential capability for autonomous robots, enabling object detection and decision-making in the presence of failing or uncertain inputs. While recent fusion methods excel in normal environmental conditions, these approaches fail in adverse weather, e.g., heavy fog, snow, or obstructions due to soiling. We introduce a novel multi-sensor fusion approach tailored to adverse weather conditions. In addition to fusing RGB and LiDAR sensors, which are employed in recent autonomous driving literature, our sensor fusion stack is also capable of learning from NIR gated camera and radar modalities to tackle low light and inclement weather. 

We fuse multimodal sensor data through attentive, depth-based blending schemes, with learned refinement on the Bird's Eye View (BEV) plane to combine image and range features effectively. Our detections are predicted by a transformer decoder that weighs modalities based on distance and visibility. We demonstrate that our method improves the reliability of multimodal sensor fusion in autonomous vehicles under challenging weather conditions, bridging the gap between ideal conditions and real-world edge cases. Our approach improves average precision by $17.2\,AP$ compared to the next best method for vulnerable pedestrians in long distances and challenging foggy scenes. Our project page is available \href{https://light.princeton.edu/samfusion/}{here}$^{1}$.

\end{abstract}

\section{Introduction}
Autonomous vehicles rely on multi-modal perception systems with  sensors such as LiDAR \cite{chen2023voxelnext,yin2021center,wang2020infofocus,lang2019pointpillars}, camera \cite{wang2021fcos3d,wang2022detr3d,grauer2014active}, and radar \cite{meyer2019automotive}, combining distinct modalities with complementary weaknesses and strengths to enable safe autonomous driving. Recent work \cite{bijelic2020seeing, xie2023sparsefusion,liu2023bevfusion,sindagi2019mvx,chen2023futr3d,cramnet,yang2022deepinteraction} combines input from these diverse sensors to enhance environment perception with accurate localization and classification of objects in captured street scenes. As such, these systems benefit from the accuracy of LiDAR depth \cite{xie2023sparsefusion}, the robustness of radar \cite{cramnet,nabati2021centerfusion}, and the dense semantic information of cameras \cite{liu2023bevfusion,sindagi2019mvx,chen2023futr3d}. Although fusion is crucial for downstream classification and localization tasks, as was shown in \cite{bijelic2020seeing, broedermann2023hrfuser,hwang2022cramnet}, when sensors fail, special care is required to achieve better results with fusion than with single camera networks.
Examples of fusion strategies include physically-inspired entropy-driven fusion, as proposed in \cite{bijelic2020seeing}, and learned attention fusion as seen in \cite{broedermann2023hrfuser}. The most effective 3D object detection methods often utilize a Bird's-Eye-View (BEV) representation, either by concatenating modality-specific feature maps \cite{liu2023bevfusion,liang2022bevfusion} or by employing multiple attention-based modules to enhance BEV features \cite{ge2023metabev,bai2022transfusion}. However, the robustness of these techniques is typically validated only on datasets collected under favorable weather conditions \cite{Geiger2013IJRR,caesar2020nuscenes}, and they have not been proven effective against adverse weather-related disturbances, such as asymmetric degradation in LiDAR point clouds \cite{bijelic2018benchmark}. This vulnerability is largely attributed to the reliance on a unimodal query generator, and dependence on LiDAR-based depth projections \cite{yang2022deepinteraction}, which can lead to network failures in the absence of reliable LiDAR data.

\begin{figure}[t!]
    \centering
    \includegraphics[width=0.92\columnwidth]{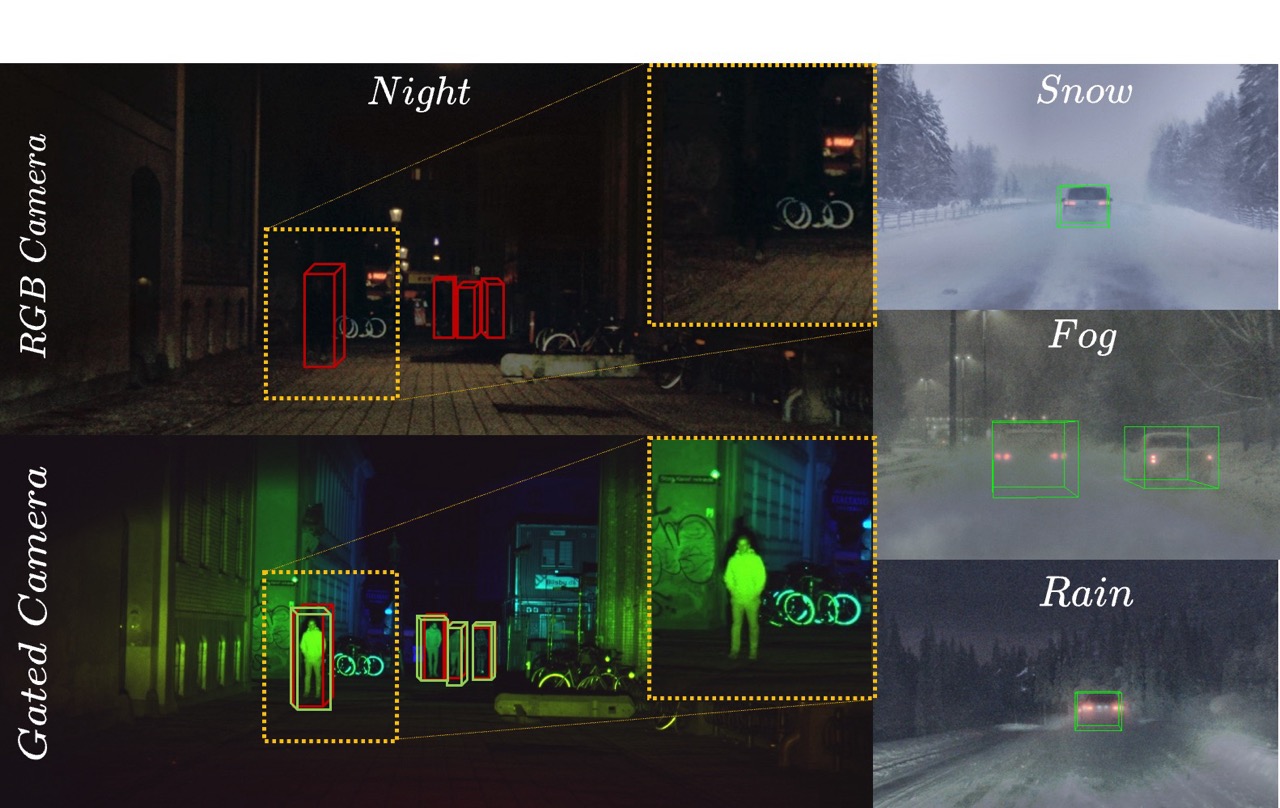}
    \caption{In this work, we present SAMFusion, a multimodal fusion approach combining gated NIR, RGB color-imaging, LiDAR, and radar point clouds for object detection in challenging adverse weather conditions. The qualitative results above show beneficial low light detection capabilities due to the gated camera as well as example detections from our proposed approach in night-time, snowy and foggy conditions, which are achieved through attentive blending of features and multimodal querying. We depict ground truth bounding boxes in red, and predictions in green.}
    \label{fig:abstractFigure}
\end{figure}

Recent advancements in gated imaging technology offer a promising alternative to conventional imaging modalities, and were explored in \cite{grauer2014active, bijelic2018benchmarking, julca2021gated3d, gated2gated, walzgatedstereo}. This work demonstrates the capability of gated cameras to actively eliminate backscatter\cite{bijelic2018benchmarking}, provide accurate depth \cite{gated2gated, walzgatedstereo}, and achieve high signal-to-noise ratios (SNR) in adverse scenarios such as night-time, fog, snowy or rainy conditions, all due to their active gated scene illumination. We will therefore use gated cameras in addition to more conventional camera, LiDAR and radar data to further increase robustness. 

In summary, we tackle the challenge of robust object detection in inclement weather by addressing two key problems in sensor fusion: modality projection quality and robustness against sensor distortions in adverse weather. To this end, we propose a sensor-adaptive multi-modal fusion method -- SAMFusion. We introduce a novel encoder structure with a depth-guided camera-LiDAR transformation and additional early fusion between both camera modalities, incorporating distance-wise precise cross-modal projections. Additionally, we introduce a novel multi-modal, distance-based query generation approach to avoid relying solely on the LiDAR modality to generate detection proposals, as in \cite{bai2022transfusion,yang2022deepinteraction}. 

\noindent Specifically, we make the following contributions:
\begin{itemize}
\setlength{\parskip}{0pt} 
\setlength{\itemsep}{-2pt plus -2pt}
    \item We propose a novel transformer-based multi-modal sensor fusion approach, improving object detection in the presence of severe sensor degradation. \\
    \item We introduce an encoder architecture combining early camera fusion, depth-based cross-modal transformation, and adaptive blending in conjunction with learned distance-weighted multimodal decoder proposals to increase the reliability of object detection across lighting and weather conditions.\\
    \item We design a transformer decoder that aggregates multimodal information in BEV through multimodal proposal initialization.\\
    \item We validate the method on automotive adverse weather scenes~\cite{bijelic2018benchmark} and improve 3D-AP, especially for the pedestrian class by more than \textbf{\unit[17.2]{AP}} in dense fog and  \textbf{\unit[15.62]{AP}} in heavy snow on the most challenging distance category from {\unit[50]{m}-\unit[80]{m}} relative to the state of the art.
\end{itemize}

\section{Related Work}
\PAR{3D Object Detection.}
The task of 3D object detection evolved from 2D object detection, requiring the prediction of 3D-bounding boxes (bboxes) and orientations of objects \cite{geiger2012we,liang2019multi,qi2018frustum,yang20203dssd,lang2019pointpillars}. Unimodal LiDAR methods, such as \cite{lang2019pointpillars,zhou2018voxelnet}, have been explored to leverage the depth accuracy of the LiDAR sensor to predict 3D bboxes based on LiDAR point clouds. Point-based methods \cite{qi2018frustum,qi2017pointnet++,shi2019pointrcnn,yang20203dssd} therefore generate detections from raw point cloud features. Other methods group LiDAR points into 3D voxels \cite{chen2022focal,chen2023largekernel3d} or pillars \cite{wang2020pillar,yin2021center}. Voxel and point-based methods can also be chained together, such as in \cite{wu2023pv,shi2020pv,shi2020points}, which implement additional refinement steps to improve 3D object detection performance based on region of interest pooling~\cite{ren2015faster,he2017mask}.
Camera-based methods were investigated in~\cite{liu2020smoke,wang2021fcos3d,liu2023monocular,liu2023petrv2}, which work in the image space itself. However, camera data has proven to be a good candidate for fusion with LiDAR, as the former can be mapped to a BEV representation, and the latter natively lives in the BEV space. Therefore, the camera representation space has since evolved from camera coordinates \cite{liu2020smoke,wang2021fcos3d} to joint multi-view setups and predicted BEV representations \cite{huang2021bevdet,li2023bevdepth}, improving 3D detection accuracy.

\PAR{Multi-modal Sensor Fusion.}
While a common BEV map is not necessarily the default choice, several multi-modal sensor fusion approaches have incorporated semantic camera information to enrich individual LiDAR points, as described in \cite{vora2020pointpainting,wang2021pointaugmenting,yin2021multimodal}. Subsequent studies, such as \cite{yoo20203d,xu2021fusionpainting}, have investigated how to extract detailed information from camera data for LiDAR point clouds, which is heavily dependent on the quality of projection and was further refined by \cite{yin2021multimodal}. These approaches introduced virtual 3D camera points to provide a more dense environmental context for enhancing sparse point clouds at long distances. Li et al. \cite{li2022unifying} extended this approach by integrating deformable attention \cite{zhu2020deformable} to create a unified representation of both modalities in the 3D voxel space.

Recently, another line of research operating in the BEV space has shown remarkable effectiveness. This approach fuses features that are aggregated in a reference frame (e.g., the LiDAR BEV perspective) and then processed by task decoders performing various perception tasks such as 3D object detection \cite{wang2023unitr,huang2024leveraging,jiao2023msmdfusion,yan2023cross,cai2023bevfusion4d, BEVFormer, liu2023petrv2}, lane estimation \cite{HDMapnet, BEVSegformer, liu2023petrv2}, tracking \cite{BEVPlanning}, semantic segmentation \cite{liu2023bevfusion, BEVFormer, liu2023petrv2}, and planning \cite{BEVPlanning}. Such a framework supports multitasking and multimodal models that benefit from the additional supervision and regularization provided by these configurations. However, even the most recent BEV representation approaches \cite{liu2023bevfusion, ku2018defense} still face challenges in projecting detailed camera features into the BEV world coordinate system and preventing error propagation in the case of sensor distortions. 

\PAR{Sensor Fusion in Adverse Weather.}
In this work, we specifically aim to tackle the degradation of individual sensors under adverse weather conditions, which drastically reduces object detection performance as shown previously in \cite{wang2023unibev, mirza2021robustness, cramnet, sun2020towards, uricar2019desoiling}. Multi-modal sensor fusion emerged as a viable approach to achieve robustness under these scenarios ~\cite{diaz2022ithaca365, mirza2021robustness, zhang2015visual, bijelic2020seeing, baumann2024cr3dt, broedermann2023hrfuser}. 
In detail, \cite{hwang2022cramnet,chen2023futr3d,lin2024rcbevdet, baumann2024cr3dt} fuse the camera modality with radar information, while \cite{bijelic2020seeing,broedermann2023hrfuser} introduce additional sensing modalities and exploit novel, physically-grounded fusion techniques. However, these only allow for the prediction of 2D object detections. Our approach projects to a common BEV plane, with attention-based feature fusion and the incorporation of dense depth to allow 
 for more performant 3D object detection.

\section{SAMFusion}
\begin{figure*}[t!]
     \centering
     \begin{subfigure}[b]{0.24\textwidth}
         \centering
         \includegraphics[height=6cm, trim={0 0 0 0}, clip]{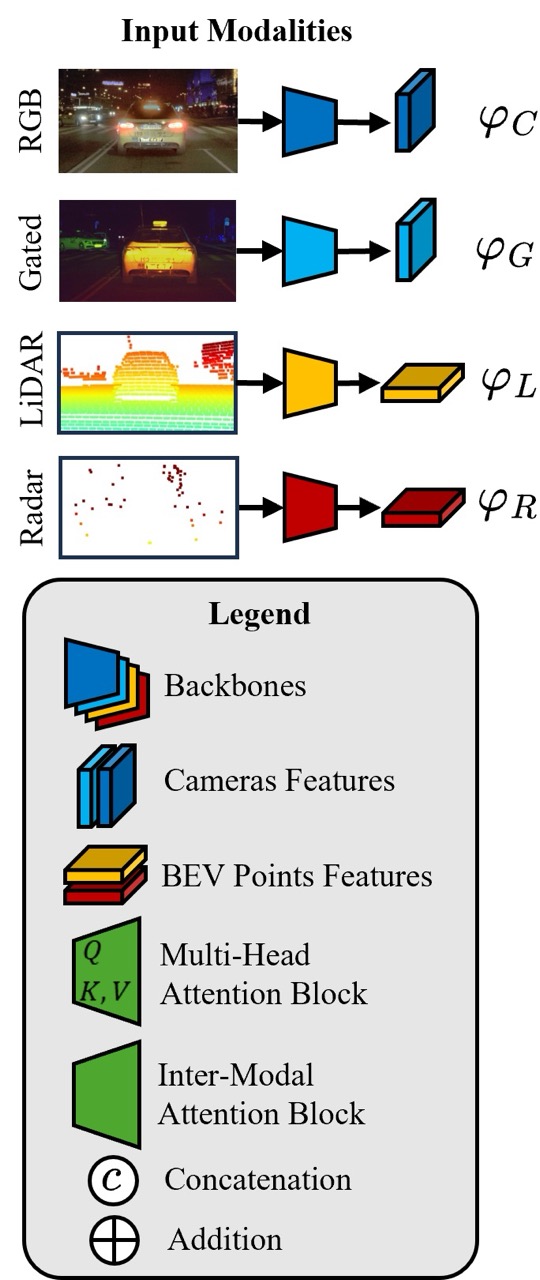}
         \caption{Backbones, legend}
         \label{Fig:SAM_legend}
     \end{subfigure}
     \hfill
     \begin{subfigure}[b]{0.4\textwidth}
         \centering
         \includegraphics[height=6cm]{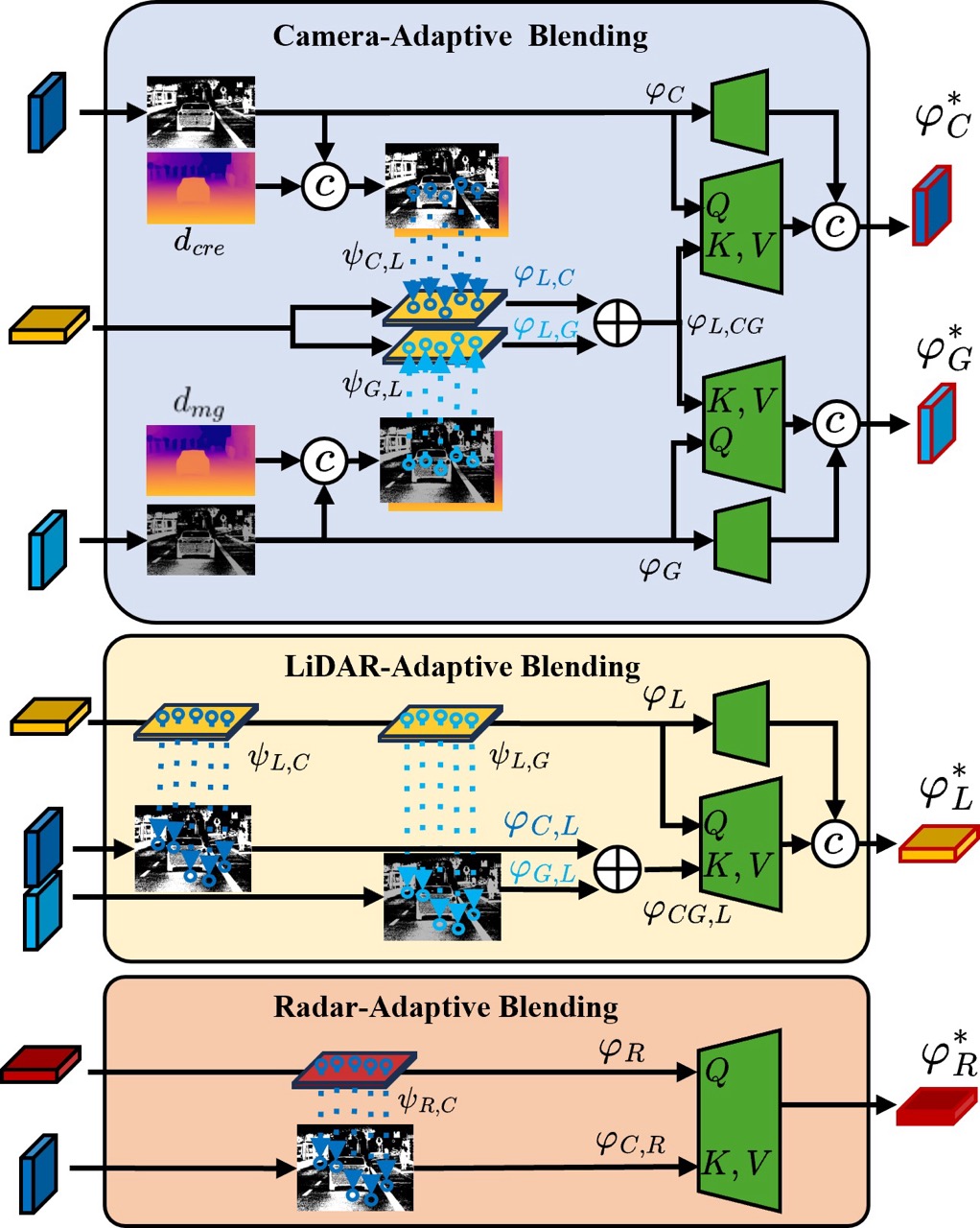}
         \caption{SAM encoder}
         \label{Fig:SAM_encoder}
     \end{subfigure}
     \hfill
     \begin{subfigure}[b]{0.34\textwidth}
         \centering
         \includegraphics[height=6cm, trim={0 0 0 0}, clip]{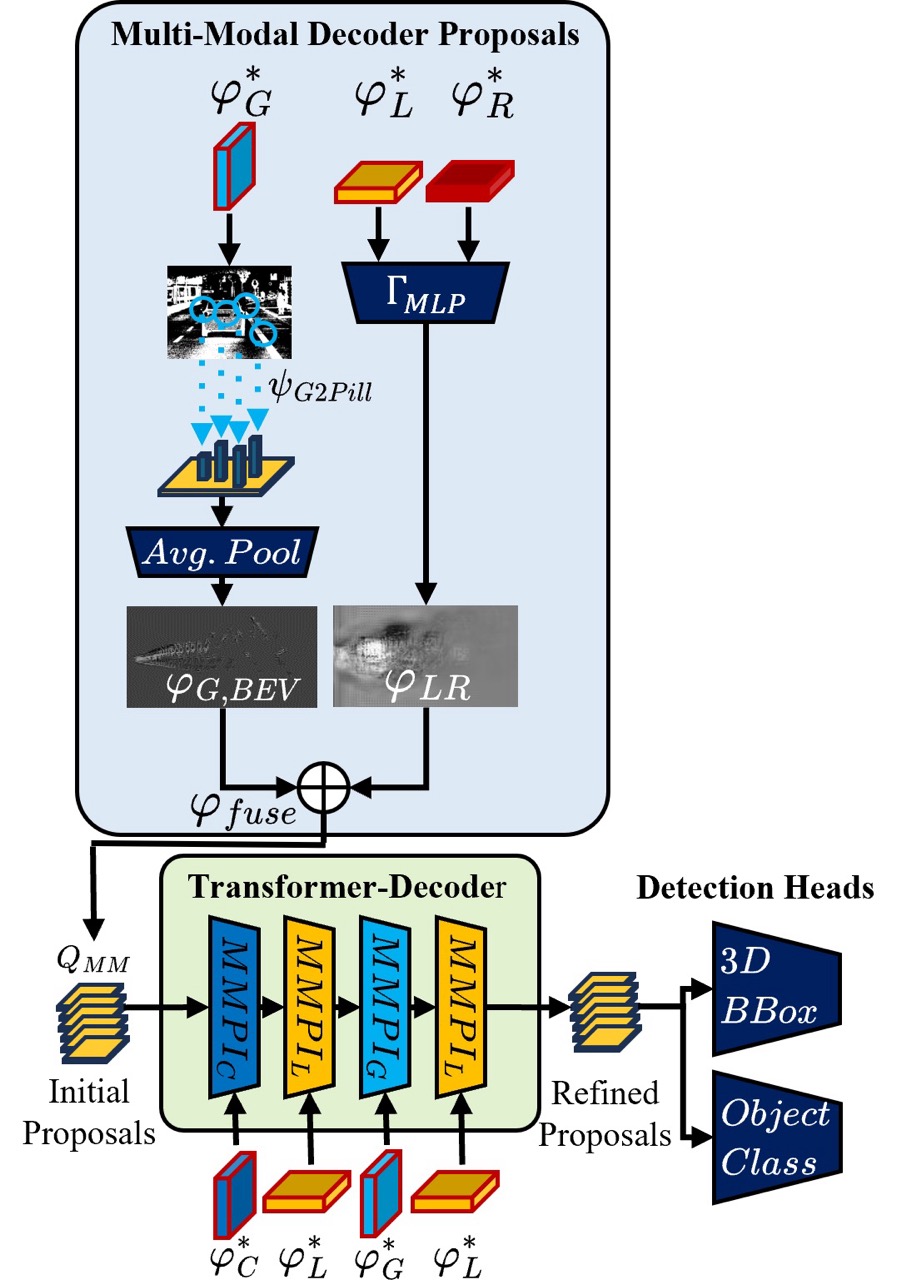}
         \caption{SAM decoder}
         \label{Fig:SAM_decoder}
     \end{subfigure}
    \caption{SAMFusion Architecture. \ref{Fig:SAM_legend} First, we extract features from each modality. \ref{Fig:SAM_encoder} Then, we refine them, fusing modalities through attention and depth-based blending. \ref{Fig:SAM_decoder} Finally, refined gated and range (LiDAR and radar) features are agglomerated in BEV, and are combined in a weighted manner that is aware of distance and weather, before being refined further and sent to detection heads to produce bounding box outputs. The gated camera and radar sensors complement the high-definition RGB camera and LiDAR to better handle poor illumination and adverse weather.}
    \label{Fig:ges}
\end{figure*}

In this section, we introduce the SAMFusion architecture for multimodal 3D object detection. SAMFusion leverages the complementary strengths of LiDAR, radar, RGB, and gated cameras. Gated cameras excel in foggy and low-light conditions, while radar is effective in rain and at long distances. By integrating these sensors into a depth-based feature transformation, a multi-modal query proposal network and a decoder head, SAMFusion ensures robust and reliable 3D object detection across diverse scenarios. The architecture is illustrated in Fig.~\ref{Fig:ges}.

The inputs - RGB/gated camera, LiDAR, radar - are transformed into features through their respective feature extractors \ref{Fig:SAM_legend}. These features are blended in the Multi-Modal encoder \ref{Fig:SAM_encoder} in an attentive fashion, and are combined with camera-specific feature maps to produce enriched features $\varphi^*$ - we refer to this as ``early fusion''.

Features $\varphi^*$ are now passed to the multi-modal decoder proposal module \ref{Fig:SAM_decoder} where they are refined with another level of fusion in the Bird's Eye View representation to combine the image features (gated camera) and the range features (LiDAR, radar) in an adaptive, distance-weighted fashion for initial object proposals. Additionally, the enriched features $\varphi^*$ are sent to the transformer decoder that refines the initial object proposals to attentively produce detection outputs. The decoder proposals include optimizations to adaptively weight distance through a learned weighting scheme that is aware of the physical properties of ranging sensors while fusing with the information-dense camera modality.

% ------------------------------------------------------------------------------------------------------------------------------------------------------------------------------------------------------------------------------------------------
% ------------------------------------------------------------------------------------------------------------------------------------------------------------------------------------------------------------------------------------------------
\subsection{Cross-Modal Adaptive Blending}\label{2camEncoder}
This section describes the early attention fusion schemes of individual sensor features. An illustration of the methodology is shown in Fig.~\ref{Fig:SAM_encoder}.

In the SAMFusion encoder, early attention fusion integrates information from different modalities. To achieve this, we first create a weighted context from the features of the primary modality, which aligns with the features of the secondary modality. This context (key) is then queried with data from the second modality (query), resulting in a rich mix of aligned features.

Our early fusion approach supports queries from both camera and LiDAR modalities, creating two parallel instances of pair-wise (query, key) attentive fusion. In "Camera-Adaptive Blending," queries from RGB and gated cameras are compared against weighted LiDAR context samples (RGB camera against Sampled LiDAR and gated camera against Sampled LiDAR). This blending accounts for objects visible in one modality but not in the other. Similarly, in "LiDAR-Adaptive Blending," LiDAR queries are scored with sampled weighted camera context features blended across RGB and gated images (LiDAR against Sampled camera).

Finally, we refine radar features in a similar fashion, where the radar proposals are scored with weighted context provided from the RGB camera.

% ------------------------------------------------------------------------------------------------------------------------------------------------------------------------------------------------------------------------------------------------
\PAR{Camera-Adaptive Blending.}\label{CGLI} 
In this module, we use attention to score the camera features $\varphi_{C}, \varphi_{G}$ (query) against the weighted context $\varphi_{L,CG}$ (keys, values) derived from the LiDAR modality. To generate such a context, we gather LiDAR BEV features $\varphi_{L}$ corresponding to the camera features. We note that the LiDAR feature encoder outputs are available in the form of a BEV image. Therefore, we transform all the camera pixels $(u, v)$ onto the LiDAR coordinate frame. In order to achieve this, we need pixel-wise depth $\textbf{d}(u,v)$ for each camera feature coordinate. In Fig. \ref{Fig:SAM_encoder} we denote the concatenation with the symbol \textcircled{$c$} that assigns the corresponding depth to each pixel.

Together with depth, we use known camera intrinsics and extrinsics (with respect to LiDAR) to lift image points into the 3D $(x, y, z)$ LiDAR coordinate space. In our setup, we compute depth differently for RGB and gated cameras. For RGB cameras, we use stereo RGB pairs from the dataset and predict depth utilizing \cite{li2022practical}, while for gated cameras, the depth ($d_{MG}$) is attained from a mono-RGB method \cite{ranftl2021vision}, which is fine-tuned on the gated camera data following \cite{walz2023gated}.

The projection - $\psi_{C,L}$ for RGB camera, $\psi_{G,L}$ for gated camera, $\psi_{C;G,L}$ - is attained by lifting the pixels into a point cloud using
\begin{equation}
\begin{cases}
      z = \textbf{d}(u,v), \\
      x = (u - C_x) \times z/f_x, \\
      y = (v - C_y) \times z/f_y, 
    \end{cases}\
\label{eq:L2C}
\end{equation}
where $(fx,fy)$ are the horizontal and vertical focal lengths of the camera and $(Cx , Cy )$ is the pixel location corresponding to the camera center, and then applying a change of frame of reference to bring the 3D points into the LiDAR coordinate frame.

The reprojected 3D camera points $(x, y, z)$ are then squashed along the height coordinate $y$ onto the LiDAR BEV grid. Further, we resolve the discretization of the LiDAR feature map $\varphi_L(x, z)$ by bilinear interpolation of the corresponding BEV coordinates. Subsequently, the found correspondences are used to enrich each 3D camera point $(x,y,z)$ with extracted LiDAR features $\varphi_L$, which are backprojected into the camera image and paired with image features prior to scoring with attention. Through this procedure, for each RGB and gated camera pixel $\varphi_C(u, v)$ and $\varphi_G(u, v)$ we obtain corresponding LiDAR feature points $\varphi_{L, C}(u, v)$ and $\varphi_{L,G}(u, v)$.

Finally, these two independent weighted LiDAR contexts are blended together to get a composite representation $\varphi_{L,CG}$ that is aware of both camera modalities. This composition is obtained by summing up the two feature maps, where we drop the positional dependence in $\varphi_{L,C}(u,v)$ and $\varphi_{L,G}(u, v)$ for notational convenience:
\begin{equation}
\label{eq:mlpencoder}
\begin{aligned}
     \varphi_{L,CG} = \varphi_{L,C}&\oplus \varphi_{L,G},
\end{aligned}
\end{equation} 
where $\oplus$ is the element-wise addition operation.

The described process is introduced to integrate detailed camera-specific information into $\varphi_{L,CG}$, avoiding the case when either modality fails due to reduced visibility of the sensors in adverse lighting conditions. 

Having obtained the associated LiDAR feature points to compare with, we integrate cross-modal attention to learn enriched modality-specific feature maps, including object features from the LiDAR modality that can be occluded in the camera frames due to the physical position of the sensors. We carry out an attention computation between the respective camera and LiDAR modalities $(\varphi_C, \varphi_{L,CG})$ and $(\varphi_G, \varphi_{L,CG})$ to produce the final enriched camera-specific feature maps $\varphi_C^*$ and $\varphi_G^*$, to guide the decoder object proposals. We write the cross-modal attentive blending equation with LiDAR (key, value) $\varphi_{L,CG}$, abbreviating the extracted RGB and gated features $\varphi_C, \varphi_G$ as $\varphi_{C;G}$ and the enriched maps $\varphi_C^*, \varphi_G^*$ as $\varphi_{C;G}^*$, as 
\begin{equation}
 \varphi_{C;G}^* = \sum_{\varphi_{L,CG}\in J_s} softmax\biggl(\frac{\varphi_{C;G}\,\varphi_{L,CG}^{T}}{\sqrt{d}}\biggr)\varphi_{L,CG}.
\label{eq:cross_modal_camera}
\end{equation}

The attention computation is performed over a local window $J_s$ around the sampled point $(i, j)$, with a window size of $k$ and a softmax normalization factor of $d$, representing the dimensionality of the point cloud features.

We note that, besides the cross-modal attention mechanism, we execute intra-modal-attention in parallel on the queried modality, described by 
\begin{equation}
 \varphi_{C;G}^* = \sum_{\varphi_{C;G}\in J_s} softmax\biggl(\frac{\varphi_{C;G}\, \varphi_{C;G}^{T}}{\sqrt{d}}\biggr)\varphi_{C;G}.
\label{eq:intra_modal_camera}
\end{equation} Afterwards, $\varphi_{C;G}^*$ feature maps, cross-modal-attention and intra-modal-attention results are fused with a learned weighting scheme (independently for RGB $\varphi_{C}$ and gated $\varphi_{G}$).

% ------------------------------------------------------------------------------------------------------------------------------------------------------------------------------------------------------------------------------------------------
\PAR{LiDAR-Adaptive Blending.}\label{lidarinteraction} In this module, we blend LiDAR features $\varphi_L$ with a weighted context from RGB and gated camera features $\varphi_{CG, L}$ using attention, with LiDAR features serving as queries and camera features as keys and values. Unlike camera-adaptive blending, depth is inherently included in the LiDAR BEV features $\varphi_L(x_L, z_L)$. Therefore, before projecting into the camera feature map, we assign the LiDAR points $(x_L, y_L, z_L)$ to columns at the respective feature map grid positions $(x_L, z_L)$.

Furthermore, the 3D LiDAR features $\varphi_L(x_L, y_L, z_L)$ are mapped onto the corresponding 2D image points $(u_{C;G, L}, v_{C;G, L})$ by projection, analogous to Eq. \ref{eq:L2C}, through the $\psi_{L, C;G}$ LiDAR-to-camera (RGB; gated) projection matrix. The camera features corresponding to relevant LiDAR feature coordinates $(u_{C;G, L}, v_{C;G, L})$ are acquired by sampling from the image modalities through bilinear interpolation. 

Next, we blend the LiDAR-aware sampled image features from the two camera modalities
\begin{equation}
\label{eq:FCGL}
\begin{aligned}
     \varphi_{CG,L} = \varphi_{C,L} \oplus \varphi_{G,L},
\end{aligned}
\end{equation}
before scoring against corresponding LiDAR queries. As before, we drop the positional dependence in $\varphi_C(u_{C,L},v_{C,L}), \varphi_G(u_{G,L}, v_{G,L})$ for notational convenience.

The enriched LiDAR feature map $\varphi_L^*$ is obtained similarly to the Camera-Adaptive-Blending in Sec. \ref{CGLI}, blending the output of the cross-modal attention between LiDAR queries and LiDAR aware image features (similarly to Eq. \ref{eq:cross_modal_camera}) to the output of the intra-modal attention over LiDAR features (as per Eq. \ref{eq:intra_modal_camera}).

% ------------------------------------------------------------------------------------------------------------------------------------------------------------------------------------------------------------------------------------------------
\PAR{Radar-Adaptive Blending.}\label{radarencoder}
In the radar branch, we rely on the same principle as for the LiDAR-Adaptive Blending described in Sec. \ref{lidarinteraction}, with the only difference being that we calculate the weighted context from the RGB camera modality only and don't perform intra-modal attention due to the sparseness of radar point clouds.

% ------------------------------------------------------------------------------------------------------------------------------------------------------------------------------------------------------------------------------------------------
% ------------------------------------------------------------------------------------------------------------------------------------------------------------------------------------------------------------------------------------------------
\subsection{Multi-Modal Decoder Proposals}\label{robustqueries}
SAMFusion generates initial object proposals $Q_{MM}$ based on a multi-modal BEV feature map with an additional learned weighting scheme, prioritizing modalities based on distance and weather. The distance weighting is encoded in the BEV-based fusion of radar and LiDAR while additional weather robustness is gained by enriching the multimodal queries with the gated modality. An example is rainy weather, where LiDAR is compromised and can be enhanced by proposals from camera and radar modalities. 

In particular $Q_{MM}$ are generated from LiDAR, radar and gated camera features. An illustration of the methodology is presented in Fig.~\ref{Fig:SAM_decoder}.

% ------------------------------------------------------------------------------------------------------------------------------------------------------------------------------------------------------------------------------------------------

\PAR{Weighted Radar And LiDAR Feature Map Fusion.}\label{distanceweighting}
We leverage distance-dependent sensor-specific ranging characteristics and employ a weighted fusion approach to combine the enriched feature maps $\varphi_L^*$ and $\varphi_R^*$ into a joint feature map $\varphi_{LR}$ described by
\begin{equation}
\varphi_{LR} = \Gamma_{MLP}(f(d, \sigma) \varphi_L^* + (1-f(d, \sigma)) \varphi_R^*)
\label{eq:mlp}
\end{equation}
where $f = \exp((-\frac{d}{2 \sigma^2})^2)$, and $d$ is the distance of each feature point from the ego veichle and $\sigma$ is a learned parameter.

The learned $\Gamma_{MLP}$ weighs LiDAR and radar features through a gaussian mask with learned variance, which amplifies LiDAR at close range and suppresses it at longer ranges to favor radar. The range is dependent on the learned guassian variance. The resulting features $\varphi_{LR}$ are thus modulated to contain LiDAR and radar, weighted by their relative importance across the ROI.

% ------------------------------------------------------------------------------------------------------------------------------------------------------------------------------------------------------------------------------------------------

% ------------------------------------------------------------------------------------------------------------------------------------------------------------------------------------------------------------------------------------------------

\PAR{Late Gated Camera Features Fusion.}\label{CameraQueries} To generate the final object proposal, our method encodes the initial proposals extracted from the gated camera. Due to the time-of-flight principle of the sensor, they encode distance within the captured intensity profiles. To encode detailed gated camera features $\varphi_G^*$ a pillar-based conditioning approach is used to transform the camera feature map into a common BEV representation matching the distance-weighted feature map $\varphi_{LR}$. The original LiDAR coordinates are transformed according to the 3D LiDAR points into the camera representation, as described in Sec.~\ref{lidarinteraction} and are used to sample camera features $\varphi_G^*$. Then, camera features are assigned to the corresponding LiDAR pillars and the feature positions in the LiDAR BEV grid are determined through average pooling, resulting in a BEV camera feature map $\varphi_{G,BEV}$.
Features $\varphi_{G,BEV}$ and $\varphi_{LR}$ are fused in an additive manner to obtain a distance-encoded weighted feature map $\varphi_{fuse}$ dependent on three modalities by conditioning the ranging sensor feature maps with corresponding gated camera features. 
Further, we apply class-dependent convolution layers onto $\varphi_{fuse}$ to extract object proposal centers based on maximum intensity values and obtain the initial object proposals $Q_{MM}$. $Q_{MM}$ sets the starting point for the decoder refinement process through Multi-Modal-Predictive-Interaction layers obtained from Yang et al. \cite{yang2022deepinteraction}. 

% ------------------------------------------------------------------------------------------------------------------------------------------------------------------------------------------------------------------------------------------------
% ------------------------------------------------------------------------------------------------------------------------------------------------------------------------------------------------------------------------------------------------
\subsection{Training}
The SAMFusion architecture, designed as a transformer network, follows the learning methodology of Carion et al.~\cite{carion2020end} and Bai et al.~\cite{bai2022transfusion}. It first matches labels to predictions using Hungarian loss \cite{HungarianAlgo}, then minimizes a loss composed of a weighted sum for classification (Cross-Entropy), regression, and IoU. Detailed loss formulations are provided in the supplemental material.

% ------------------------------------------------------------------------------------------------------------------------------------------------------------------------------------------------------------------------------------------------
% ------------------------------------------------------------------------------------------------------------------------------------------------------------------------------------------------------------------------------------------------
\subsection{Implementation}
We implement SAMFusion in PyTorch \cite{paszke2017automatic} and the open-source library MMDetection3D \cite{mmdet3d2020}. We initialize the camera branch with a ResNet-50 \cite{he2016deep} backbone and pretrained Cascade Mask R-CNN \cite{cai2019cascade} weights. The original RGB and gated camera images are scaled with center-based cropping to [800,400] to reduce computational 
cost. 
We define the voxels to be 0.075\,m deep, 0.075\,m wide and 0.2\,m high. We restrict the LiDAR and radar point clouds to (0\,m, 100\,m) in range and to (-40\,m, 40\,m) in width. The height range is set to (-3\,m, 1\,m) and (-0.2\,m, 0.4\,m) for LiDAR and radar respectively.
We implement four stacked transformer decoder layers, guided by RGB, gated camera, and LiDAR modalities with 200 initial multi-modal proposals. We train all models for 12 epochs in an end-to-end manner with a batch size of 4 on NVIDIA V100 GPUs. Refer to the supplemental material for hyperparameter and training settings on the SeeingThroughFog dataset~\cite{bijelic2020seeing} as well as a full latency comparison against multi-modal sensor fusion methods, proving the real-time capabilities of our approach.

\section{Experiments}
In this section, we present experiments validating the design choices of SAMFusion. Subsection~\ref{sec:useddatasets} introduces the metrics and datasets, Subsection~\ref{sec:ablationExperiments} presents ablations of the individual contributions and Subsection~\ref{sec:assesment} showcases comparisons against existing state-of-the-art uni- and multi-modal 3D detection methods on day, night, foggy and snowy scenarios.

\begin{table*}[t!]
	\caption{Evaluation of SAMFusions detection performance measured in AP and compared to State-of-the-Art mono- and multi-modal methods based on the car and pedestrian classes on the SeeingThoughFog \cite{bijelic2020seeing} test set.}
	\centering
\begin{subtable}[c]{0.9\columnwidth}
	\setlength{\tabcolsep}{0.4em}
	\centering
	\vspace*{-4pt}
	\resizebox{\columnwidth}{!}{
    \begin{tabular}{l|c|cccccc|cccccc}
        \multicolumn{14}{c}{Average Precision for \textit{Pedestrian} class} \\
		\midrule

		\multirow{3}{*}{\textbf{Method}} & \multirow{3}{*}{\textbf{Modality}} & \multicolumn{6}{c|}{\textbf{Day}} & \multicolumn{6}{c}{\textbf{Night}} \\
  
		& & \multicolumn{3}{c}{\textbf{3D object detection}} & \multicolumn{3}{c|}{\textbf{BEV detection}} & \multicolumn{3}{c}{\textbf{3D object detection}} & \multicolumn{3}{c}{\textbf{BEV detection}} \\
  
		& & 0-30m & 30-50m & 50-80m & 0-30m & 30-50m & 50-80m & 0-30m & 30-50m & 50-80m & 0-30m & 30-50m & 50-80m\\
		\midrule
		\midrule

		\textsc{M3D-RPN} \cite{brazil2019m3d} & C & 26.20 & 14.50 & 9.84 & 30.68 & 17.47 & 10.07 & 25.09 & 6.43 & 2.07 & 26.42 & 7.69 & 2.74 \\

		\textsc{PatchNet} \cite{ma2020rethinking} & G & 32.88 & 18.05 & 5.62 & 39.45 & 20.27 & 9.77  & 15.37 & 13.37 & 6.75 & 21.60 & 18.15 & 8.46 \\
		\textsc{Gated3D} \cite{julca2021gated3d} & G & 50.94 & 20.59 & 14.14 & 53.26 & 22.15 & 16.51 & 48.53 & 23.99 & 14.98 & 49.82 & 25.57 & 15.46 \\
		\textsc{Stereo-RCNN} \cite{li2019stereo} & S & 48.58 & 23.26 & 7.77 & 50.11 & 25.10 & 8.38 & 46.09 & 21.63 & 11.57 & 47.58 & 25.47 & 11.84 \\	
		
        \textsc{SECOND} \cite{second} & L & 70.75 & 51.81 & 19.34 & 71.05 & 52.51 & 20.28 & 69.04 & 48.09 & 14.56 & 70.51 & 49.23 & 15.32 \\
         
		\textsc{MVXNet} \cite{sindagi2019mvx} & CL & 74.51 & 61.69 & \underline{29.78} & 74.88 & 62.63 & \underline{30.54} &  \underline{74.15} & 55.66 & \underline{23.19} &  \underline{74.42} & 55.90 & \underline{23.58} \\        
        
		\textsc{BEVFusion} \cite{liang2022bevfusion} & CL & 64.25 & 57.91 & 8.86 & 64.76 & 59.41 & 8.86 & 65.78 & 52.91 & 7.25 & 66.25 & 54.40 & 7.27 \\              
        
        \textsc{DeepInteraction} \cite{yang2022deepinteraction} & CL &  \underline{78.01} & \underline{66.59} & 28.55 &  \underline{77.98} & \underline{66.67} & 28.54 & 71.98 & \underline{61.10} & 20.53 & 71.96 & \underline{61.29} & 20.72 \\

        \textsc{SparseFusion} \cite{xie2023sparsefusion} &  CL & 68.27  & 60.18 &  16.89 & 68.18 &  60.32 &  16.92 & 61.11  &   57.09 & 12.67 & 61.21 & 57.24 & 12.66  \\
        
        \textsc{\textbf{SAMFusion}} &  \textbf{CGLR} &  \textbf{80.09}& \textbf{70.97}& \textbf{40.16}& \textbf{79.97}& \textbf{70.99}& \textbf{40.35}& \textbf{75.49}& \textbf{67.59}& \textbf{27.14}& \textbf{75.49}& \textbf{67.56}& \textbf{27.16}\\

	\end{tabular}}
	\label{tab:3dgated_object_detection_results_pedestrian}
	%\vspace*{4pt}
	\end{subtable}%
    \hspace{1em}
    \vspace{1em}
    \\
    \begin{subtable}[c]{0.9\columnwidth}
	\setlength{\tabcolsep}{0.4em}
	\centering
	\resizebox{\columnwidth}{!}{
 	\begin{tabular}{l|c|cccccc|cccccc}
        \multicolumn{14}{c}{Average Precision for \textit{Car} class} \\
		\midrule

		\multirow{3}{*}{\textbf{Method}} & \multirow{3}{*}{\textbf{Modality}} & \multicolumn{6}{c|}{\textbf{Day}} & \multicolumn{6}{c}{\textbf{Night}} \\
  
		& & \multicolumn{3}{c}{\textbf{3D object detection}} & \multicolumn{3}{c|}{\textbf{BEV detection}} & \multicolumn{3}{c}{\textbf{3D object detection}} & \multicolumn{3}{c}{\textbf{BEV detection}} \\
  
		& & 0-30m & 30-50m & 50-80m & 0-30m & 30-50m & 50-80m & 0-30m & 30-50m & 50-80m & 0-30m & 30-50m & 50-80m\\
		\midrule
		\midrule

		\textsc{M3D-RPN} \cite{brazil2019m3d} & C & 53.21 & 13.26 & 10.52 & 60.80 & 16.16 & 10.52 & 51.18 & 20.76 & 2.73 & 52.53 & 21.39 & 2.74 \\

		\textsc{PatchNet} \cite{ma2020rethinking} & G & 23.91 & 10.86 & 7.34 & 24.87 & 11.33 & 7.84 & 23.74 & 16.79 & 7.16 & 25.15 & 17.76 & 8.29 \\
		\textsc{Gated3D} \cite{julca2021gated3d} & G & 52.15 & 28.31 & 14.85 & 52.31 & 29.26 & 15.02 & 51.42 & 25.73 & 12.97 & 53.37 & 29.13 & 13.12 \\
		\textsc{Stereo-RCNN} \cite{li2019stereo} & S & 54.17 & 17.16 & 6.17 & 57.92 & 17.69 & 6.26 & 47.36 & 17.21 & 13.02 & 53.81 & 18.34 & 13.08 \\
        \textsc{SECOND} \cite{second}& L & 95.68 & 81.90 & 46.81 & 95.70 & 82.18 & 47.55 & 98.01 & 84.10 & \underline{48.53} & 98.03 & 84.23 & \underline{50.39} \\

		\textsc{MVXNet} \cite{sindagi2019mvx} & CL & 96.29 & 84.09 & 50.35 & 96.30 & 84.09 & \underline{51.83} & 96.36 & 85.99 & \textbf{49.79} & 96.36 & 86.06 & \textbf{51.17} \\        
        
		\textsc{BEVFusion} \cite{liang2022bevfusion}& CL & 95.30 & 86.86 & 11.43 & 95.43 & 87.38 & 11.24 & 93.89 & 84.84 & 12.17 & 93.95 & 85.31 & 12.48 \\        

        \textsc{DeepInteraction} \cite{yang2022deepinteraction}& CL & 97.12 & \underline{87.95} & \textbf{51.84} & 97.13 & \underline{88.47} & \textbf{51.99} & \underline{98.31} & \underline{88.09} & 46.83 & \underline{98.31} &\underline{ 88.11} & 46.87 \\

        \textsc{SparseFusion} \cite{xie2023sparsefusion} & CL & \textbf{97.47} & 88.10 & 31.02 & \textbf{97.49} & 88.26 & 31.11 & 96.12 & 86.49 & 27.99 & 96.13 & 86.51 & 28.01 \\

        \textsc{\textbf{SAMFusion}} &  \textbf{CGLR} &  \underline{97.25} &  \textbf{89.50} & \underline{50.68} & \underline{97.26}& \textbf{89.69}&50.80& \textbf{98.77}& \textbf{88.91}&44.40 & \textbf{98.82}& \textbf{89.16}&45.46\\

	\end{tabular}}
	\label{tab:3dgated_object_detection_results_car}
	%\vspace*{4pt}
	\end{subtable}
	%\vspace*{-0pt}
	\label{tab:3dgated_object_detection_results}
\end{table*}

\subsection{Dataset And Evaluation Metrics}\label{sec:useddatasets}
This section describes the evaluation of SAMFusion on the SeeingThroughFog dataset \cite{bijelic2020seeing}, consisting of 12,997 annotated samples in adverse weather conditions, covering night, fog, and snowy scenarios in Northern Europe. Following \cite{julca2021gated3d}, we divide the dataset into 10,046 samples for training, 1,000 for validation, and 1,941 for testing. The test split is further divided into 1,046 daytime and 895 nighttime samples, with respective weather splits.
Additionally, we provide results for our evaluations on the NuScenes dataset \cite{caesar2020nuscenes} in the supplemental material. 

\PAR{Evaluation Metrics.} Object detection performance is evaluated according to the metrics specified in the KITTI evaluation framework \cite{geiger2012we}, including 3D-AP and BEV-AP for the passenger car and pedestrian class. We incorporate 40 recall positions \cite{simonelli2019disentangling} for the AP calculation. To match the predictions and ground truth we apply intersection over union (IoU) \cite{chen2017multi} with an IoU of 0.2 for passenger cars and 0.1 for pedestrians. Further,  we follow \cite{yang2018pixor} and report results according to respective distance bins. 

\begin{figure*}[t!]
	\centering
	\includegraphics[width = \textwidth]{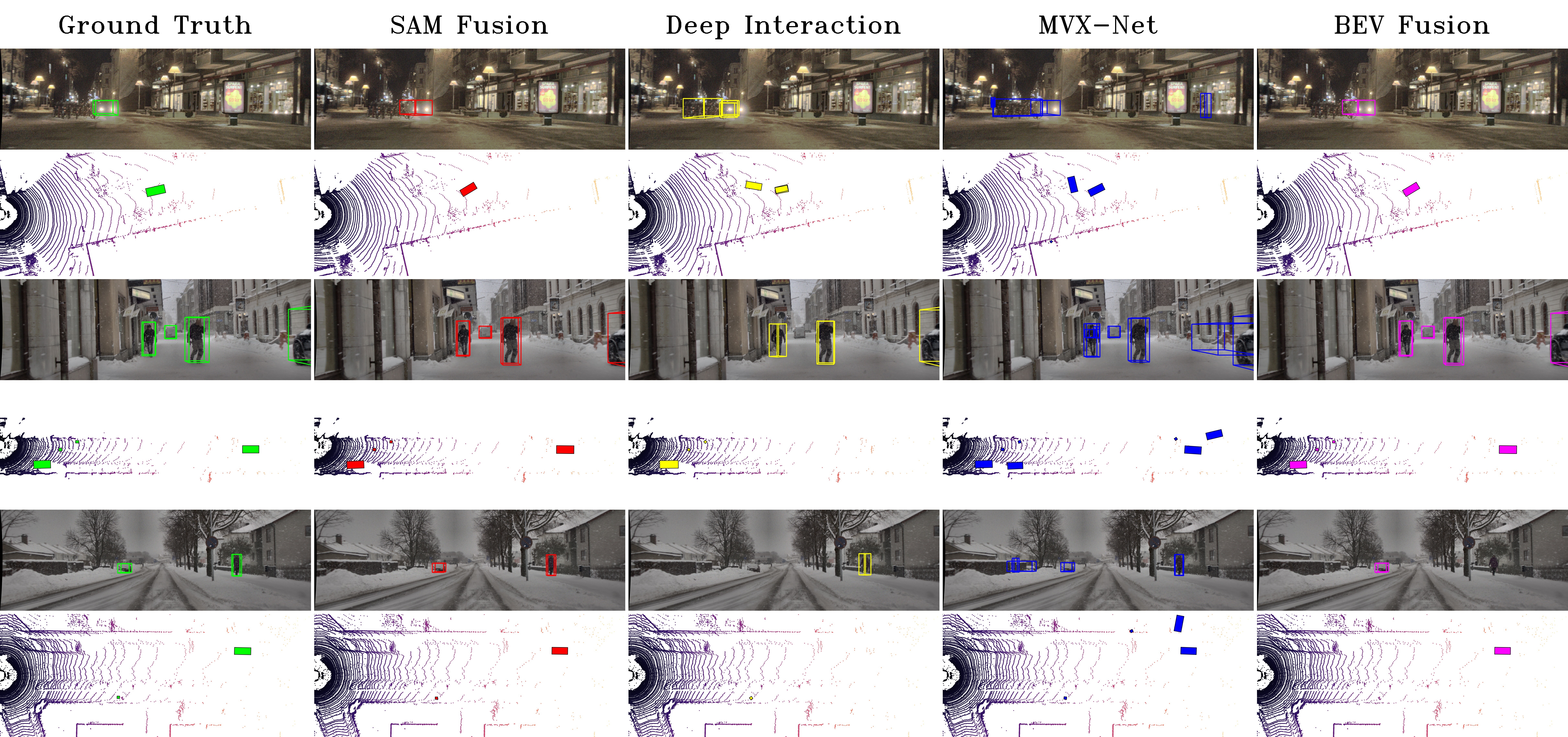}
	\caption{Qualitative results on 3D Object detection in adverse weather compared to state-of-the-art multi-modal sensor fusion methods and ground truth (GT). While all methods perform well in the daytime setting, SAMFusion outperforms other reference methods in adverse and low light conditions (rain, snow, fog, twilight, night). In rainy and snowy settings, other methods show missing (BEVFusion) or spurious (MVXNet, DeepInteraction) detections, especially for the pedestrian class. In twilight and night, the effects are more pronounced, with missing and erroneous detections in most objects. Moreover, we see SAMFusion excel with far-away objects and pedestrian detection.}
	\label{Fig:Detections}
\end{figure*}
\begin{figure*}[t!]
	\centering
	\includegraphics[width = \textwidth]{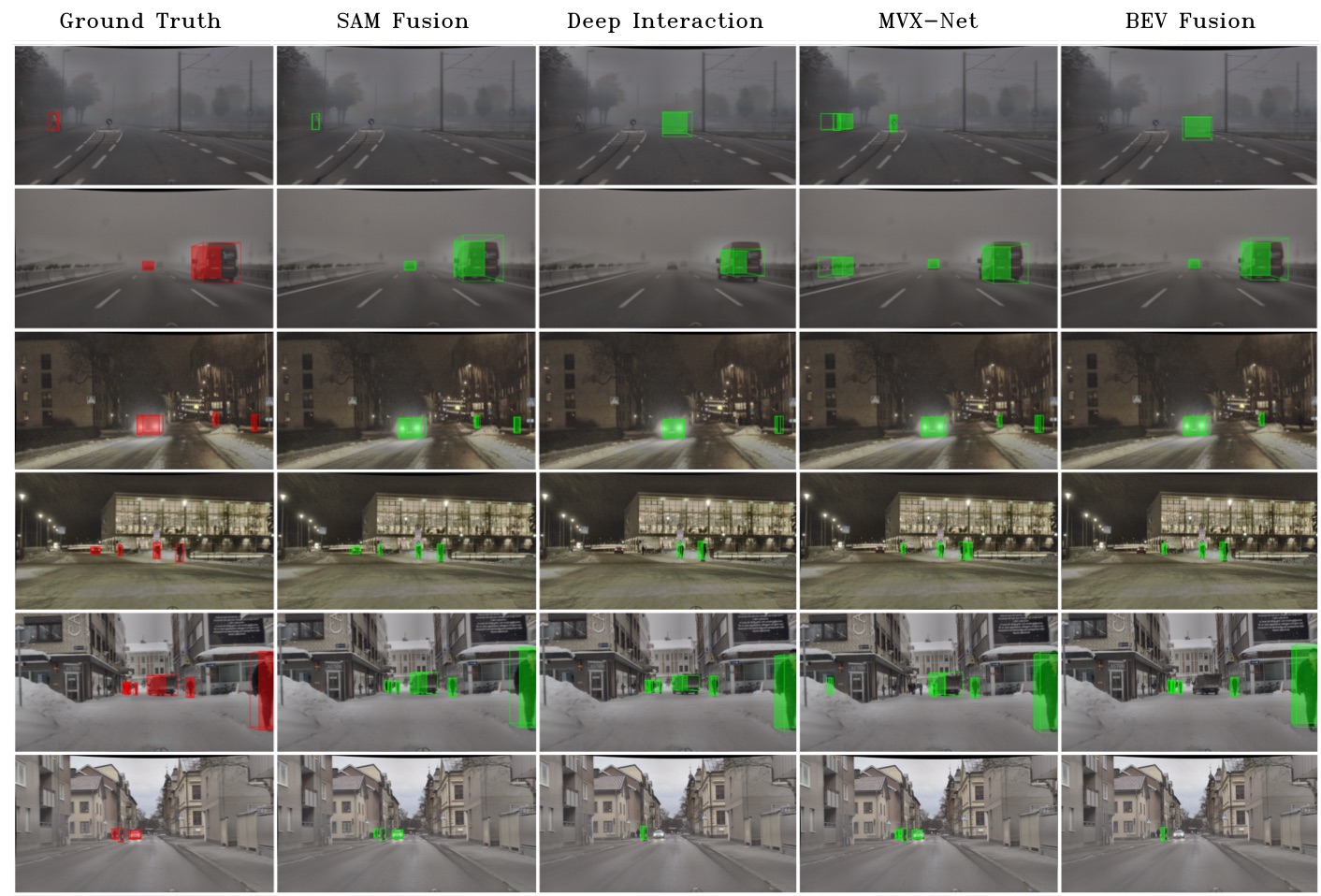}
    \caption{We show qualitative results on different sequences (rows) of the proposed method and reference approaches (columns). 
    On the left the ground truth is illustrated with red bounding boxes, followed by the proposed SAMFusion approach, BEVFusion \cite{liang2022bevfusion}, MVXNet \cite{sindagi2019mvx} and DeepInteraction \cite{yang2022deepinteraction}.}
	\label{Fig:Detections_2}
\end{figure*}

\subsection{Ablation Experiments}\label{sec:ablationExperiments}
In this subsection, we validate our methodological contributions shown in Table~\ref{tab:differentmod} and Table~\ref{tab:samemod}.

Table~\ref{tab:differentmod} explores ablations with varying numbers of input modalities using the SAMFusion architecture. Configurations include single camera-LiDAR (CL), gated-LiDAR (GL), camera-LiDAR-radar (CLR), gated-LiDAR-radar (GLR), and camera-gated-LiDAR-radar (CLGR) inputs. These methods utilize queries based on LiDAR and radar data with learned distance weightings. We focus our results on the pedestrian class at extended distances, where detection is most challenging due to sparse LiDAR points. The outcomes underscore the benefits of integrating additional modalities, particularly noticeable during both day and night conditions.

Performance comparisons between single camera modalities with passive RGB and active gated imaging (GL and CL) show distinct advantages under different lighting conditions. In daylight, the inclusion of RGB color information in CL provides a performance boost of 2.85 AP-points within the 50\,m to 80\,m range. Conversely, at night, the superior SNR of active illumination in GL enhances detection, yielding improvements of +1.08\,AP in mid-range and +3.45\,AP in long-range distances. Integrating both camera technologies in the CGL configuration leverages the strengths of each, delivering enhanced performance across day and night settings. The addition of radar data further amplifies overall performance, although the absence of the gated camera slightly diminishes night-time efficacy.

The optimal results manifest when all four modalities (CGLR) are used, capitalizing on the unique strengths of each sensor to bolster the architecture’s resilience across diverse lighting and adverse weather conditions. This configuration also benefits from leveraging proposals generated from all involved modalities.

Further, in Table~\ref{tab:samemod}, we extend our validation to assess the impact of our fusion technique beyond mere modality integration. We investigate the efficacy of depth-based transformations, weighted BEV maps, and various modal proposal strategies. The incremental inclusion of these methodological enhancements correlates with notable performance improvements, indicating that simply stacking modalities is insufficient for maximizing results. For instance, incorporating multi-modal proposals elevates night-time pedestrian detection by 15.2\% over solely point cloud-based proposals. Additionally, our distance-aware weighting mechanism, $\Gamma_{MLP}$, further boosts detection capabilities by up to 20.7\%. Notably, proposals utilizing gated imaging data yield a larger improvement margin than those based on color data, due to their inherent distance encoding, which facilitates superior geometrical localization.

\begin{table*}[!tb]
	\caption{
     We measure the individual method contributions on the most difficult pedestrian class.
     Table \ref{tab:differentmod} ablates the addition of new modalities as input and in the proposal generation. We observe that adding beneficial sensor modalities improves pedestrian detection reliability, especially in low light conditions. Fusing both cameras in the adaptive blending module boosts overall detection quality of small objects due to detailed camera specific feature maps with significant information content in far distances.
    Table \ref{tab:samemod} ablates the proposal modality configurations and the depth-based transformations in the encoder and the learned $\Gamma$-weighting for LiDAR-radar-fusion. Object detection results are evaluated based on the 3D AP metric explicitly for the pedestrian class and the most relevant far distance from 50-80m.
    } 
    \vspace*{-12pt}
	\centering
    \begin{subtable}[t]{0.48\columnwidth}
	\setlength{\tabcolsep}{0.4em}
	\centering
	%\vspace*{-4pt}
    \subcaption{Ablation of Input Modality configurations.}
    %\vspace*{-2pt}
	\resizebox{\columnwidth}{!}{
    \begin{tabular}{l|c|c|cc|cc}
		& \multicolumn{1}{c|}{\textbf{Input}} & \multicolumn{1}{c|}{\textbf{Proposal}} &\multicolumn{2}{c|}{\textbf{Day}} & \multicolumn{2}{c}{\textbf{Night}} \\
	  &\multicolumn{1}{c|}{\textbf{Modality}}&\multicolumn{1}{c|}{\textbf{Modality}}&\multicolumn{2}{c|}{\textbf{3D object detection}} & \multicolumn{2}{c}{\textbf{3D object detection}} \\
		& & & 30-50m & 50-80m  & 30-50m & 50-80m\\
		\midrule
		\midrule
  \multirow{10}{*}{\rotatebox[origin=l]{90}{\parbox[c]{0.6cm}{\centering \textbf{\textsc{Ablation}}}}}
          & CL & L & 66.59 & 28.55  & 61.10 & 20.80\\
          & GL & L & 65.59 & 26.89 & 63.25 & 22.11\\
         & CGL & L & 66.88 & 28.94 & 64.17 & 22.34\\
        
        & CLR  & LR & 69.06 &35.02 & 65.97 & 20.95 \\	
        
         & GLR  & LR & 69.52 & 32.17 & 67.05 &24.40 \\

        & CGLR  & LR & \underline{69.98} & \underline{35.60} & \underline{67.22} & \underline{26.85} \\
        
        & CGLR  & GLR & \textbf{70.99} & \textbf{40.16} & \textbf{67.56} & \textbf{27.14} \\
	\end{tabular}}
	\label{tab:differentmod}
	%\vspace*{4pt}
	\end{subtable}%
	%\vspace*{4pt}
    \hspace{1em}
    \begin{subtable}[t]{.48\linewidth}
	\setlength{\tabcolsep}{0.4em}
	\centering
    %\hspace{-3pt}
	%\vspace*{-4pt}
    \subcaption{Ablation of SAMFusion components.}
    %\vspace*{-2pt}
	\resizebox{\linewidth}{!}{
		\begin{tabular}{l|c|c|cccccccc}
			%\toprule
			& \textbf{Input}& \textbf{Depth-based}&\multicolumn{4}{c}{\textbf{Proposal}}& &   \multicolumn{1}{c}{\textbf{Day}} &\multicolumn{1}{c}{\textbf{Night}} &  \\
            &\textbf{Modality}&\textbf{Transformation}&\multicolumn{4}{c}{\textbf{Modality}}&\textbf{$\Gamma_{MLP}$}& \\
			&  & & C & G & R & L &&  50-80m & 50-80m & \\
			\midrule
            \midrule
			\multirow{9}{*}{\rotatebox[origin=l]{90}{\parbox[c]{0.6cm}{\centering \textbf{\textsc{Ablation}}}}}
	       & CGLR & \xmark & \xmark & \xmark & \xmark & \cmark &  \xmark   &28.94& 22.34\\
   & CGLR & \xmark & \xmark & \xmark & \cmark & \cmark  &  \xmark    & 29.48 &  23.02 \\
            & CGLR & \cmark & \xmark & \xmark & \cmark & \cmark &  \xmark  & 29.49  & 24.01  \\
            & CGLR & \cmark & \xmark & \xmark & \cmark & \cmark & \cmark  & 35.60  & \underline{26.85} \\
            & CGLR & \cmark & \cmark & \xmark & \cmark & \cmark &  \cmark & \underline{36.19} &22.79\\
            & CGLR & \cmark & \xmark & \cmark & \cmark & \cmark &  \cmark & {\textbf{40.16}}&{\textbf{27.14}} \\

	       \end{tabular}}
    	\label{tab:samemod}
	\end{subtable}
\end{table*}

\begin{table*}[!tb]
	\caption{
    Detection performance of SAMFusion measured in AP compared to multi-modal methods in challenging weather conditions, evaluated on the car and pedestrian classes of weather test splits from~\cite{bijelic2020seeing}. We achieve significant performance increases shown in the last row of each Table.
    } 
    \vspace*{-10pt}
	\centering
    \begin{subtable}{.98\linewidth}
    	\setlength{\tabcolsep}{0.4em}
    	\centering
    	\vspace*{-4pt}
    	\resizebox{\columnwidth}{!}{
        \begin{tabular}{l|c|ccc|ccc|ccc|ccc}
            \multicolumn{2}{c}{} & \multicolumn{6}{c|}{Average Precision for \textit{Pedestrian} class} & \multicolumn{6}{c}{Average Precision for \textit{Car} class} \\

    		\multirow{3}{*}{\textbf{Method}} & \multirow{3}{*}{\textbf{Modality}} & \multicolumn{3}{c|}{\textbf{Snow}} & \multicolumn{3}{c|}{\textbf{Fog}} & \multicolumn{3}{c|}{\textbf{Snow}} & \multicolumn{3}{c}{\textbf{Fog}}\\
      
    		& & \multicolumn{3}{c|}{\textbf{3D Object Detection}} & \multicolumn{3}{c|}{\textbf{3D Object Detection}} & \multicolumn{3}{c|}{\textbf{3D Object Detection}} & \multicolumn{3}{c}{\textbf{3D Object Detection}}\\
      
    		& & 0-30m & 30-50m & 50-80m & 0-30m & 30-50m & 50-80m & 0-30m & 30-50m & 50-80m & 0-30m & 30-50m & 50-80m \\
    		\midrule
    		\midrule
             
    		\textsc{MVXNet} \cite{sindagi2019mvx} & CL &\underline{76.23}&59.73&\underline{25.83}&73.89 &50.98&16.73
            &95.82&86.02&50.28&92.81 &84.62&\underline{52.30}\\        
            
    		\textsc{BEVFusion} \cite{liang2022bevfusion} & CL &71.12&62.61&10.01&76.24&58.04&8.61
            &92.55&89.74&10.79&92.20&84.04&13.97\\              
            
            \textsc{DeepInteraction} \cite{yang2022deepinteraction} & CL & 72.91 &  57.56 &  18.38  & 66.62 & 50.32 & 10.64 
            & 95.36  & 82.05   &\underline{56.21} &95.44  &83.55&49.30\\
    
            \textsc{SparseFusion} \cite{xie2023sparsefusion} &  CL  & 73.33 & \underline{66.84} & 19.87 & \underline{79.25} & \underline{58.39} & \underline{17.05}
            & \underline{96.79} & \underline{91.35} & 32.11 & \underline{95.81} & \underline{87.71} & 25.16 \\
    
            \textsc{\textbf{SAMFusion}} &  \textbf{CGLR} & \textbf{87.44} & \textbf{80.51} & \textbf{41.45} &\textbf{83.18}   &\textbf{66.96}&\textbf{34.31}
            & \textbf{97.36} &  \textbf{93.06}  & \textbf{56.22} &\textbf{96.50}  &\textbf{92.41}&\textbf{52.99} \\
            
     	\hline
             \multicolumn{1}{c}{\textbf{Improvement in AP}}& & \textbf{+11.2} & \textbf{+13.6} & \textbf{+15.62}&\textbf{+3.9} & \textbf{+8.5} & \textbf{+17.2}
             & \textbf{+0.5} & \textbf{+1.7} & \textbf{+0.01} &\textbf{+0.7} & \textbf{+4.6} & \textbf{+0.7}\\
    	\end{tabular}}
	\end{subtable}%
	\vspace{1em}
    \label{tab:3dgated_object_detection_results_snow_fog}
\end{table*} 

\subsection{Assessment}\label{sec:assesment}
We compare SAMFusion against nine state-of-the-art methods, including one monocular camera 3D object detection method \cite{brazil2019m3d}, two gated camera methods \cite{ma2020rethinking,julca2021gated3d}, one stereo camera approach \cite{li2019stereo}, one LiDAR approach \cite{second}, and four LiDAR-RGB fusion methods \cite{sindagi2019mvx,liang2022bevfusion,yang2022deepinteraction,xie2023sparsefusion}. The results are summarized in Table \ref{tab:3dgated_object_detection_results} and further qualitative assessments are presented in Figure~\ref{Fig:Detections} and ~\ref{Fig:Detections_2}, with reported detections in both BEV and perspective view.

SAMFusion outperforms all state-of-the-art multi-modal methods in pedestrian detection under adverse weather and varying lighting conditions. Particularly in the far distance range of $50\,m$ to $80\,m$, SAMFusion achieves margins of up to $34.85\%$ during the day and $17.03\%$ during the night for 3D pedestrian detection. Additionally, pedestrian detection performance increases in mid-range distances by  $10.6\%$. These improvements can be attributed to the enhanced visibility at night arising from additional active sensors, but also to their effective incorporation through a multi-modal distance-based weighting scheme.

Car detection improves slightly. This is due to labeling bias in the car category for 3D annotations, which prioritize precision over completeness. Objects with fewer than five LiDAR points were marked as "don't care", making it difficult to measure improvements in such challenging cases. For pedestrians, a different strategy focusing on completeness was employed, thereby providing a greater amount of challenging ground truth labels not available for the car category.

\PAR{Adverse Weather Evaluation.}
Table~\ref{tab:3dgated_object_detection_results_snow_fog} validates the proposed method in adverse weather, like snow and fog. State of the art LiDAR-RGB methods struggle with reduced visibility and back-scatter in adverse weather, causing such fusion approaches to perform significantly worse than in clear conditions, despite the relatively simple scene configurations. Relative to these baselines, SAMFusion achieves improvements of up to $13.6\,AP$ ($20.4\%$ relative) for pedestrians at mid-range and $15.62\,AP$ ($60.51\%$ relative) at long-range compared to the second-best (LiDAR and RGB) method in snowy scenes. In foggy scenes SAMFusion achieves high margins of up to $17.2\,AP$ ($101.2\%$ relative) for pedestrians. For the car class in foggy conditions, it achieves improvements of up to $4.6\,AP$ ($5.2\%$ relative).

Detection performance in adverse weather correlates with scene difficulty. The relative improvement in performance compared to Table \ref{tab:3dgated_object_detection_results} can be explained by the reduced number of road users in these weather splits simplifying the general task at hand as less people participate in road traffic.

\section{Conclusion}
We propose SAMFusion, a multi-modal adaptive sensor fusion method for robust 3D object detection in adverse weather for autonomous driving. Our approach enhances the conventional camera-LiDAR perception stack with gated camera and radar sensors, significantly improving performance in low-light and adverse weather scenarios, particularly for detecting narrow-profiled and vulnerable road users. SAMFusion employs depth-based adaptive blending of sensing modalities in conjunction with a learned multi-modal, distance-weighted decoder-query mechanism that leverages sensor-specific visibility over distance. We validate our method on the challenging SeeingThroughFog dataset \cite{bijelic2020seeing}, achieving an improvement of $17.2\,AP$ points for pedestrians in dense fog and $15.62\,AP$ points in heavy snow at long range.
Future work will incorporate additional tasks such as planning and propagating uncertainty in adverse weather for improved decision-making and trajectory planning, further enhancing the robustness and effectiveness of autonomous driving systems in challenging conditions.

\PAR{Acknowledgments} This work was supported by the AI-SEE project with funding from the FFG, BMBF, and NRC-IRA. Felix Heide was supported by an NSF CAREER Award (2047359), a Packard Foundation
Fellowship, a Sloan Research Fellowship, a Sony Young Faculty Award, a Project X Innovation Award, and an Amazon Science Research Award.
Further the authors would like to thank Stefanie Walz, Samuel Brucker, Anush Kumar, Fathemeh Azimi and David Borts.

\bibliographystyle{splncs04}
\bibliography{main}
\end{document}